RESEARCH ARTICLE                                    OPEN ACCESS

# Brain Tumor Detection Based On Symmetry Information

Narkhede Sachin G, Prof. Vaishali Khairnar

*Abstract*
Advances in computing technology have allowed researchers across many fields of endeavor to collect and maintain vast amounts of observational statistical data such as clinical data, biological patient data, data regarding access of web sites, financial data, and the like.
This paper addresses some of the challenging issues on brain magnetic resonance (MR) image tumor segmentation caused by the weak correlation between magnetic resonance imaging (MRI) intensity and anatomical meaning. With the objective of utilizing more meaningful information to improve brain tumor segmentation, an approach which employs bilateral symmetry information as an additional feature for segmentation is proposed. This is motivated by potential performance improvement in the general automatic brain tumor segmentation systems which are important for many medical and scientific applications.

## I. Introduction

In Image processing, edge information is the main clue in image segmentation. But, unfortunately, it can't get a better result in analysis the content of images without combining other information. So, many researchers combine edge information with some other methods to improve the effect of segmentation [1] [2] [3].

Nowadays, the X-ray or magnetic resonance images have became two irreplaceable tools for tumours detecting in human brain and other parts of human body [4][5]. Although MRI is more expensive than the X-ray inspection, the development of its applications becomes faster because of the MR inspection does less harm to human than X-ray's.

Segmentation of medical images has the significant advantage that interesting characteristics are well known up to analysis the states of symptoms. The segmentation of brain tissue in the magnetic resonance imaging is also very important for detecting the existence and outlines of tumours. But, the overlapping intensity distributions of healthy tissue, tumor, and surrounding edema makes the tumor segmentation become a kind of work full of challenge. We make use of symmetry character of brain MRI to obtain better effect of segmentation. Our goal is to detect the position and boundary of tumours automatically based on the symmetry information of MRI.

## II. Literature Survey

In most of time, the edge and contrast of X-ray or MR image are weakened, which leads to produce degraded image. So, in the processing for this kind of medic image the first stage is to improve the quality of images. Many researchers have developed some effective algorithms about it [4] [5] [6].

After the quality of image been improved, the next step is to select the interesting objects or special areas from the images, which is often called segmentation. Many techniques have been applied on it. In this paper, we mainly discuss the brain tumor segmentation from MRI. For now, there are also some very useful algorithms, such as mixture Gaussian model for the global intensity distribution [7], statistical classification, texture analysis, neural networks and elastically fitting boundaries, etc. An automatic segmentation of MR images of normal brains by statistical classification, using an atlas prior for initialization and also for geometric constraints. Even through, Brain tumours is difficult to be modeled by shapes due to overlapping intensities with normal tissue and/or significant size. Although a fully automatic method for segmenting MR images presenting tumor and edema structures is proposed in, but they are all time consuming in some degree. As we know, symmetry is an important clue in image perception. If a group of objects exhibit symmetry, it is more likely that they are related in some degree. So, many researchers have been done on the detection of symmetries in images and shapes.

I developed an algorithm based on bilateral symmetry information of brain MRI. Our purpose is to detect the tumor of brain automatically. Compared with other automatic segmentation methods, more effective the system model was constructed and less time was consumed.

## III. Problem Statement

Brain tumors are a heterogeneous group of central nervous system neoplasms that arise within or adjacent to the brain. Moreover, the location of the tumor within the brain has a profound effect on the patient's symptoms, surgical therapeutic options, and the likelihood of obtaining a definitive diagnosis. The location of the tumor in the brain also markedly alters the risk of neurological toxicities that alter the patient's quality of life.

At present, brain tumors are detected by imaging only after the onset of neurological symptoms. No early detection strategies are in use, even in individuals known to be at risk for specific types of brain tumors by virtue of their genetic makeup. Current histopathological classification systems, which are based on the tumor's presumed cell of origin, have been in place for nearly a century and were updated by the World Health Organization in 1999. Although satisfactory in many respects, they do not allow accurate prediction of tumor behaviour in the individual patient, nor do they guide therapeutic decision-making as precisely as patients and physicians would hope and need. Current imaging techniques provide meticulous anatomical delineation and are the principal tools for establishing that neurological symptoms are the consequence of a brain tumor.

There are many techniques for brain tumor detection. I have used edge detection technique for brain tumor detection.

## IV. The Proposed Mechanism
### 4.1. Our algorithm
Our algorithm composes of two steps. The first is to define the bilateral symmetrical axis. The second is to detect the region of brain tumor.

### 4.1.1 Symmetry axis defining
The first step of our algorithm is mainly based on symmetry character of brain MRI. The bilateral symmetry character is very obviously in four MR Images of brain presented in Figure.1.

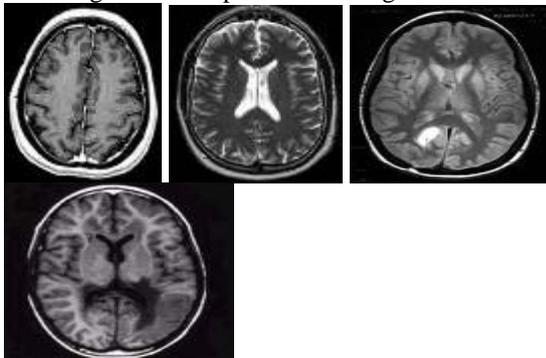

**Figure.1.** The bilateral symmetry character is very obviously.

If without tumor in the brain or the size of tumor is very small, the symmetry axis can be defined with a straight line $x = k, (y >= 0)$, which separates the image into two bilateral symmetry parts, show as Figure.2.

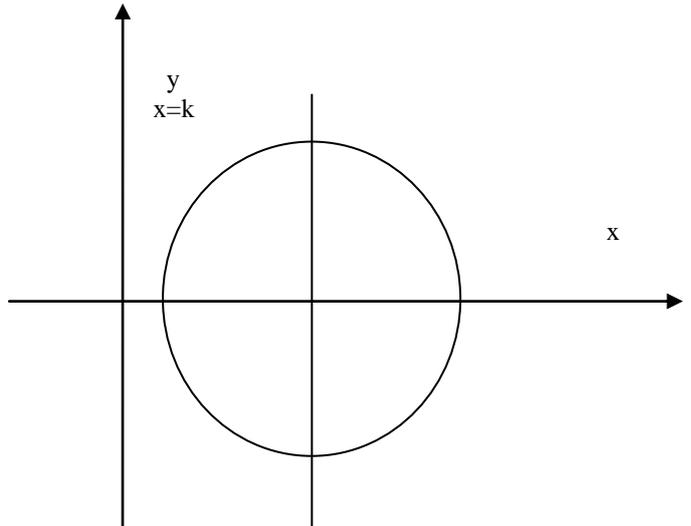

**Figure.2.** The bilateral symmetry axis is defined with a straight line.

This kind of symmetry is not very strictly. And, compared with normal brain MRI, the symmetry characteristic is distorted for the existing of brain tumor, such as the circumstance shown in Figure.3.a.

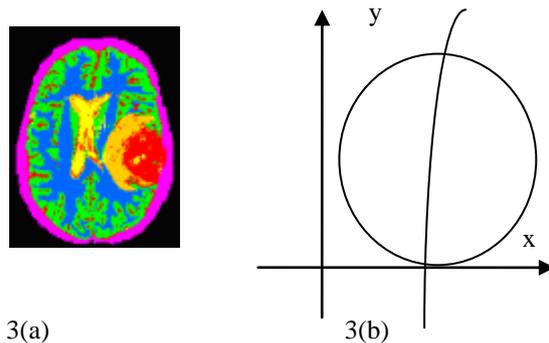

3(a)                3(b)

**Figure.3.** the symmetry axis can't be defined with a straight line in the brain MRI with tumors, so a curve line is more convenient to describe it.

For more convenient to describing symmetry axis, a curve line ( $y = f(x), x > 0, y > 0$ ) is defined, which is shown in Figure.3.b. From the edge map, the edge point set $Pe$ can be obtained. And then, we calculate the edge-centroid $Gi$ of every line according to equation (1).

$$G_i = \frac{1}{k} \sum_{j=1}^{k} P_{i,j}, \quad P_{i,j} \in Pe$$

where, $Gi$ is the abscissa of centroid in the $i$ th line, $k$ is the edge point number in the $i$ th line, whose abscissas are $P_{i,1}...P_{i,k}$. So, based on the edge-centroids, we can use the least square method to get the symmetry curve line $y$ approximatively.

## V. Methodology Used
There are many techniques for brain tumor detection. I have used edge detection technique for brain tumor detection. Edge-based method is by far the most common method of detecting boundaries and discontinuities in an image. The parts on which immediate changes in grey tones occur in the images are called edges. Edge detection techniques transform images to edge images benefiting from the changes of grey tones in the images.

## VI. Performance Evaluation

- If cutting of brain image gives symmetry by axis then there will not be chances of tumor this is detected by first algorithm otherwise there will be chances of tumor.
- As in others there are various steps are required to just identify whether there is tumor or not but in this it shows exact region where tumor is occurred.
- The color image is changes into gray scale image and then by reiterative processing the tumor is getting identified.
- Our purpose is to detect the tumor of brain automatically.
- Compared with other automatic segmentation methods, more effective the system model was constructed and less time was consumed.

Table 6.1: Number of detected edges

| Patient ID | Grade | Number of Detected Edges | | |
|---|---|---|---|---|
| | | Robert | Prewitt | Canny |
| 397384 | High | 5259 | 4382 | 1997 |
| 1941040 | High | 5120 | 4323 | 1836 |
| 1953042 | High | 6807 | 5757 | 2302 |
| 197906 | Low | 1491 | 649 | 317 |
| 1956041 | Low | 2509 | 1080 | 433 |
| 1943061 | Low | 2567 | 1072 | 417 |

Table 6.2: Areas of tumor

| Patient ID | Lesion | Volume of tumor areas (Pixels) | % of Damage areas |
|---|---|---|---|
| 397384 | Left Frontal Parietal | 4315 | 17.26 |
| 19410407 | Left High Parietal | 1068 | 4.27 |
| 19530428 | Left Temporal Lobe | 435 | 1.74 |
| 19790628 | Left Frontal Parietal | 1776 | 7.10 |
| 19560416 | Left Thalamus | 1060 | 4.24 |
| 19430618 | Left High Parietal | 3824 | 15.30 |

## VII. Conclusion

At first, it checks the image can be divided into symmetric axis or not. If it is divided into Symmetric part then no tumor in brain and it can be divided in curve shape then chances of tumor in human brain. However, if there is a macroscopic tumor, the symmetry characteristic will be weakened. According to the influence on the symmetry by the tumor, develop a segment algorithm to detect the tumor region automatically.